\documentclass[runningheads]{llncs}
\pdfoutput=1

\usepackage{url}
\urldef{\mails}\path|{gcharwat, wallner, woltran}@dbai.tuwien.ac.at|

\usepackage{amssymb,amsfonts,amsmath}

\usepackage{tikz,pgf}
\tikzstyle{arg}=[draw, thick, circle]
\usetikzlibrary{arrows}
\usepackage{float}
\usepackage{times}

\newcommand{\ie}{i.e.~}

\newcommand{\body}[1]{B(#1)}
\newcommand{\bodyp}[1]{B^+(#1)}
\newcommand{\bodyn}[1]{B^-(#1)}
\newcommand{\BU}{{\ensuremath{B_{\cal U}}}}
\newcommand{\commadots}[0]{,\ldots ,}
\newcommand{\derives}{\la}
\newcommand{\GP}{\ensuremath{Gr(\p)}}
\newcommand{\head}[1]{H(#1)}
\newcommand{\la}{\leftarrow}
\newcommand{\naf}{{\it not}\,}
\newcommand{\p}{\ensuremath{{\pi}}}
\newcommand{\U}{{\ensuremath{\cal U}}}
\newcommand{\UP}{\ensuremath{U_{\p}}}

\newcommand{\att}{\mathrm{defeat}}
\newcommand{\defeat}{\att}
\renewcommand{\succ}{\mathrm{succ}}
\renewcommand{\inf}{\mathrm{inf}}
\renewcommand{\sup}{\mathrm{sup}}
\renewcommand{\nsucc}{\mathrm{nsucc}}
\newcommand{\ninf}{\mathrm{ninf}}
\newcommand{\nsup}{\mathrm{nsup}}

\newcommand{\order}{<}

\newcommand{\lt}{\mathrm{lt}}

\newcounter{aspcounter} %\setcounter{aspcounter}{1}

\newenvironment{asp}[2]
{\refstepcounter{aspcounter} \setcounter{equation}{0} \label{#2} \par\noindent \begin{minipage}{\textwidth} \scriptsize  \begin{eqnarray} 
\pi_{#1} & =  \big \{ &}
{\; \big \} \end{eqnarray} \end{minipage}\\[5pt]}
\newenvironment{aspworkaround}[1]
{\refstepcounter{aspcounter} \setcounter{equation}{0} \label{#1} \par\noindent \begin{minipage}{\textwidth} \scriptsize  \begin{eqnarray} 
}
{ \end{eqnarray} \end{minipage}\\[5pt]}
\newcommand{\cl}{\mathrm{cl}}
\newcommand{\pclaim}{\mathrm{sclaim}}
\newcommand{\subformula}{\mathrm{subformula}}
\newcommand{\kb}{\mathrm{kb}}
\newcommand{\ptrue}{\mathrm{true}}
\newcommand{\pfalse}{\mathrm{false}}
\newcommand{\fs}{\mathrm{fs}}
\newcommand{\entailsclaim}{\mathrm{entails\_claim}}
\newcommand{\ismodel}{\mathrm{ismodel}}
\newcommand{\nomodel}{\mathrm{nomodel}}
\newcommand{\atom}{\mathrm{atom}}

\newcommand{\noatom}{\mathrm{noatom}}
\newcommand{\modelupto}{\mathrm{model\_upto}}
\newcommand{\hasmodel}{\mathrm{hasmodel}}

\newcommand{\fand}{\mathrm{and}}

\newcommand{\fneg}{\mathrm{neg}}

\newcommand{\fimp}{\mathrm{imp}}

\newcommand{\key}{\mathrm{key}}
\newcommand{\kentail}{\mathit{entail}}
\newcommand{\kmin}{\mathit{m}}

\newcommand{\modelcheck}{\mathrm{modelcheck}}
\newcommand{\entailment}{\mathrm{entailment}}
\newcommand{\consistent}{\mathrm{consistent}}
\newcommand{\mini}{\mathrm{minimize}}
\newcommand{\arguments}{\mathrm{arguments}}

\newcommand{\selectedA}{\mathrm{selected1}}
\newcommand{\selectedB}{\mathrm{selected2}}
\newcommand{\fsconj}{\mathrm{fs\_conj}}
\newcommand{\as}{\mathrm{as}}
\newcommand{\support}{\mathrm{support}}
\newcommand{\attack}{\mathrm{attack}}
\newcommand{\attacktype}{\mathrm{attacktype}}
\newcommand{\attsat}{\mathrm{att\_sat}}
\newcommand{\attacksp}{\mathrm{attacks}}

\newcommand{\checkDefeat}{\mathrm{checkDefeat}}

\newcommand{\checkDirectdefeat}{\mathrm{checkDirectdefeat}}
\newcommand{\directdefeat}{\mathrm{ddefeat}}

\newcommand{\knowledgebase}{\mathcal{K}}
\newcommand{\presetclaims}{\mathcal{C}}

\newcommand{\ASPVis}{ARVis}

\titlerunning{Utilizing ASP for Generating and Visualizing Argumentation Frameworks}
\authorrunning{G.~Charwat \emph{et al}\/.}

\begin{document}

\setcounter{page}{51}

\mainmatter  % start of an individual contribution	

\title{Utilizing ASP for Generating and Visualizing Argumentation Frameworks}
%\title{Vispartix: Utilizing Answer Set Programming for Generating and Visualizing Argumentation Frameworks}

%\titlerunning{Lecture Notes in Computer Science: Authors' Instructions}

\author{G\"unther Charwat \and Johannes Peter Wallner \and Stefan Woltran}

\institute{Vienna University of Technology, Institute of Information Systems 184/2,\\
Favoritenstra\ss e 9-11, 1040 Vienna, Austria\\
\mails%\\
%\url{http://www.dbai.tuwien.ac.at}%
}


%\toctitle{Lecture Notes in Computer Science}
%\tocauthor{Authors' Instructions}
\maketitle

%\todo[inline]{title too long}

%\pagestyle{empty}

\begin{abstract}
Within the area of computational models of argumentation, 
the instan\-tiation-based approach is gaining more and more
attention, not at least because meaningful input
for Dung's abstract frameworks is provided in that way.
In a nutshell, the aim of instantiation-based argumentation
is to form, from a given knowledge base, a set of arguments
and to identify the conflicts between them. The resulting
network is then evaluated by means of extension-based semantics
on an abstract level, i.e.\ on the resulting graph. While
several systems are nowadays available for the latter step,
the automation of the instantiation process itself has received 
less attention. In this work, we provide a novel approach to 
construct and visualize an argumentation framework from a given
knowledge base. The system we propose relies on Answer-Set 
Programming and follows a two-step approach. A first program 
yields the logic-based arguments as its answer-sets; a second 
program is then used to specify the relations between arguments 
based on the answer-sets of the first program. As it turns out,
this approach not only allows for a flexible and extensible tool 
for instantiation-based argumentation, but also provides a new method 
for answer-set visualization in general. 
%In fact, our new tool differs
%from previous visualization systems in the sense that its
%main aim is to support the user in analyzing the relation
%between answer sets in a graphical way, instead of a
%graphical representation of the single answer sets.
%We believe that this idea offers several interesting
%features and can be used in several applications going far
%beyond the area of argumentation, which was
%our initial motivation for this tool.
%
%
%some main points: first visualization which is driven by answer set program and main application for this is AF instantiation.
%The abstract should summarize the contents of the paper and should
%contain at least 70 and at most 150 words. It should be written using the
%\emph{abstract} environment.
%\keywords{We would like to encourage you to list your keywords within
%the abstract section}
\end{abstract}

\section{Introduction}

%\todo[inline]{Decide whether Thomas and Andi as co-authors}
% --> goes to acknowledgments

%One such application is 
Instantiation-based
argumentation \cite{CaminadaA07} is a central paradigm in nonmonotonic 
reasoning since it gives a formal handle to separate the logical and non-classical 
contents of reasoning in the presence of contradicting information.
Hereby, one starts with a knowledge base
and constructs arguments from it. Arguments typically consist of two parts, 
namely a support, which is grounded in the knowledge base and a claim derived from it. 
In \cite{BesnardH01} the process is described with an underlying propositional knowledge base 
using minimal sets of consistent support classically entailing the claim.
In a second step, conflicts between these arguments have to be identified. 
The obtained arguments and the relation between them 
yield a so-called argumentation framework \cite{Dung95}. 
This simple, yet expressive formalism is basically a 
directed graph whereby the arguments are represented via vertices and the conflicts with directed edges. 
% TBF ... more details? cites!
Argumentation frameworks are then evaluated with one of the 
numerous semantics for abstract argumentation available, resulting in potentially multiple acceptable sets 
of arguments \cite{BaroniG2009}.

Here we are only interested in the instantiation part, however, 
which received less attention wrt. realized systems. Notable exceptions are the Carneades system, which can construct arguments using heuristics \cite{Gordon2010} %, the query-based Epistemic and Practical Reasoner\footnote{\url{http://www.wietskevisser.nl/research/epr/}} 
and the recent TOAST implementation for the ASPIC+ framework \cite{SnaithR2012}. %
The reason for the lack of implementations is potentially twofold: First, due to the inherent high complexity  of the problem; already
constructing a single argument is hard for the second level of the polynomial hierarchy \cite{ParsonsWA03}.
Secondly, standard instantiation schemes for propositional knowledge bases result in infinite argumentation frameworks 
even for finite knowledge bases \cite{AmgoudBV11}. % (basically, since no restrictions on supports are assumed).
The first obstacle calls for highly expressive
languages, making 
answer-set programming 
\cite{BrewkaET11,mare-trus-99,niem-99} (ASP, for short)
a well suited candidate. For the
second obstacle, we restrict ourselves here to arguments
that have their claims coming from an a priori specified
set of formulae. 

To summarize, we aim here for 
a system which takes as input
a knowledge base as well as a set of potential claims and
produces the instantiated argumentation framework, such that
the latter can be processed by other argumentation tools, 
e.g.\ ASPARTIX \cite{EglyGW10} or CEGARTIX \cite{DvorakJWW12}. 
More specifically, our contributions are as follows:
%In fact, the main contributions of this work are as follows:
%\todo{cite ASPARTIX, CEGARTIX, ...}.
%
%As a solution, we present here a novel %ASP-based 
%method 
%called 
%\vispartix{}. Its main features are  %of the system are as follows:
%with the following main features:
\begin{itemize}
\item We provide ASP programs\footnote{\url{http://dbai.tuwien.ac.at/proj/argumentation/vispartix/}} to encode the construction of arguments 
as well as the construction of the conflicts. For the second task,  
the answer-sets of the first encoding are used as input.
%
%First, this system perfectly meets the 
%aforementioned requirements for instantiation-based 
%argumentation and in addition 
Thus we can make use of the high sophistication 
modern ASP systems have reached \cite{GebserKKOTM11,LeonePFEGPS06}. 
Moreover,  since the argument construction and conflict identification
are declaratively described via ASP code, the system is easily 
adaptable to other notions of arguments or conflicts. %\todo{maybe elaborate...}.
\item 
We present a system %\footnote{\url{http://www.dbai.tuwien.ac.at/proj/argumentation/vispartix/#ARVis}}
that, on the one hand, takes care
of passing the answer-sets from one program to another. 
On the other hand, the system uses the answer-sets of the two programs for visualization in form of a graph. In our case we obtain an argumentation framework. 
% On the other hand, 
% the system uses the results of the two programs to visualize the resulting
%graph. %, i.e.\ the resulting argumentation framework. 
Finally, this result 
can be exported and used by other systems for abstract argumentation.
\end{itemize}

As a by-product, we observed that 
%the \vispartix{} 
this method
is by no means restricted to the argumentation
domain. %\todo{rewrite sentence} 
Ultimately, it 
allows for a user-driven %(via a program) 
graph representation of the collection 
of answer-sets of a given input program, 
%as a graph, 
thus acting as a tool for ASP visualization 
in general.   
%In more detail,
The most interesting feature of the tool is
that the concrete specification for two answer-sets being
in relation is given by an ASP program itself. 
%This makes
%the tool very flexible in its usage providing different
%ways of visualization. 
%In its current version, \vispartix{} 
%can be used in combination with dlv and clasp.
%\todo{GC: some technical details needed} TBF: say some more. in particular, different ways to 
%construct and export the graph. 
%also: DLV-Wrapper?, 
%JAVA aspects (which graph 
%library used for presentation).
%
%%%%%%%%%%5
In recent years, %the area of answer-set programming
%\cite{BrewkaET11,niem-99,mare-trus-99}
ASP has benefited from the rising number of development and
visualization tools, e.g.\ %\cite{xxx} 
ASPViz \cite{CliffeVBP08}, ASPIDE \cite{FebbraroRR11}, Kara \cite{KloimuellnerOPT2011} and IDPDraw \cite{Wittocx2009}. 
% \todo{add some related visualization tools, e.g. sealion, ...}
These tools so far have focused on
%domain-tailored 
presentations of single answer-sets
of the given program. However, in certain applications it
is not only the single answer-sets which are of interest, but the relation between them. 

While visualization is a rather new research branch in ASP, 
it has gained more attention in the argumentation community, where 
dedicated visualization tools have been proposed already in 
the late 90s (e.g.\ \cite{AshleyPLA2007,bording2002dunes,EglyGW10,KaracapilidisP2001,ReedR04,SchneiderVB2007,Shum08,Verheij1998}, including Debategraph\footnote{\url{http://debategraph.org}} 
%Debateopedia\footnote{\url{http://idebate.org/}} 
and Rationale\footnote{\url{http://rationale.austhink.com/}}). % \todo{add discussion on argu-visualization tools, debategraph.org only as website?}.
Many of these support the argument construction by a user via different means, such as automated reasoning, input masks and database querying.
Compared to these systems, our approach combines the computational power
of high-sophisticated ASP systems with visualization aspects. Moreover, thanks to the 
declarative nature of the ASP encodings specifying the instantiation step, 
we believe that the strength of our approach lies in its flexibility and its expandability to new
argumentation formalisms.
%we believe that our approach is more flexible than existing systems and easily extensible to new
%argumentation formalisms.

% outline
This paper is structured as follows: We briefly introduce argumentation and ASP in Section~\ref{sec:preliminaries}. Then, in Section~\ref{sec:inst_based_argu}, we present 
ASP encodings for constructing an argumentation framework. In Section~\ref{sec:visual} we outline a novel visualization tool which is used for representing relations between answer-sets. 
A final discussion is given in Section~\ref{sec:conclusion}.

% \todo[inline]{check http://www.cs.uu.nl/groups/IS/archive/henry/aspicAF.pdf}

% \todo[inline]{references seems to be from COMMA 2012 demonstration; check aspic+ demonstrator: http://ova.computing.dundee.ac.uk/test-aspic.php}

%\paragraph{Outline.} \todo{write me or remove me}If there is space.

%%%%%%%%%%%%%%

%Short argumentation intro
%explain problem (afs relatively simple, much research in development of fast algorithms, but: 
%how to obtain afs in the first place?)

%idea: from knowledge base to arguments and abstract argumentation (e.g. Besnard and Hunter [2001])
%and visualization, automatic

%approach: 
%use of answer set programming, java for visualization

%'towards covering the overall process'

%\paragraph{Related Work.}

% what do say: 
% 1.) there exist some visualization tools for ASP
% 2.) mainly concerned with one answer set -> one visualization
% 3.) a) ASPWiz takes one problem asp module and one visualization module
%     b) IPDraw and Kara may have one module; all three have pre-defined visualization predicates
% 4.) mention related AF instantion&visualization work

\section{Preliminaries}
\label{sec:preliminaries}

We provide the necessary background from argumentation theory and ASP for this work. In particular we will explain the argumentation process 
based on argumentation frameworks \cite{Dung95} 
as well as briefly recall the concepts for disjunctive logic programs. %and modeling techniques such as 
%checking properties of all elements in a set 
%and the saturation technique applicable for solving $\SigmaP{2}$-hard problems. \todo{mention here complexity?}

\subsection{Argumentation} \label{sec:argumentation}

% argu intro
In this section we introduce formal argumentation. We start with the underlying process \cite{CaminadaA07}, which we will utilize in our context. 
The general process consists of three steps. First, given a knowledge base, arguments and their relationships 
are instantiated. After this instantiation the arguments are treated as abstract entities, without considering their concrete content. 
Secondly conflicts are resolved using appropriate semantics on the abstract instantiation and finally conclusions are drawn. 
% \todo{mention consistency}

In this work the knowledge base $\knowledgebase$ is a 
(potentially inconsistent)
set of 
propositional logic formulae. %We denote the signature of the logic with $\signature{}$ and
We construct the formulae with the usual connectives $\neg, \lor, \land, \rightarrow$, the negation, disjunction, conjunction and implication, respectively. %\todo{might note other connectives as well? xor, equivalence}
Furthermore entailment and logical equivalence of formulae is denoted by $\models$ and $\equiv$, respectively. We write formulae with lowercase Greek letters 
$\alpha, \beta, \gamma, ...$.

% \todo{JW: make all definitions inline}
% \begin{definition}%{Knowledge base}
% A knowledge base, denoted by $\knowledgebase$, is a set of propositional logic formulae.
% \end{definition}

\begin{example}
\label{ex:kb}
Consider the following simple and inconsistent example knowledge base: %$\knowledgebase = \{a,a \rightarrow b, \neg b \}$.
\small 
\begin{equation*}
\knowledgebase = \{a,a \rightarrow b, \neg b \}
\end{equation*}
\normalsize
\end{example}

The instantiation step now constructs arguments and relations among them based on the information available in $\knowledgebase$ according to \cite{BesnardH01}. The abstract 
representation we utilize for this purpose is the widely studied 
argumentation framework \cite{Dung95}. An argumentation framework (AF) is a directed graph $F = (\mathit{Args},\mathit{Att})$, with the vertices ($\mathit{Args}$) being abstract arguments and the directed edges ($\mathit{Att}$) denote attacks between them to represent conflicts. %We first introduce this simple framework and then proceed to how $\knowledgebase$ yields arguments and their relations.

% \begin{definition}\label{def:af}
% An {\em argumentation framework (AF)} is a pair $\AF=(Arg,Att)$ where $Arg$ 
% %\subseteq \U$ is a 
% is a set of arguments and $Att \subseteq Arg \times Arg$ is the attack relation.
% The pair 
% $(a,b) \in Att$ means that $a$ attacks 
% $b$. 
% \end{definition}
%For an AF $F=(B,S)$ we use $A(F)$ to refer to $B$ and $R(F)$ to refer to $S$. When clear from the context, we often write $a\in F$ (instead of $a\in A(F)$) and $(a,b)\in F$ (instead of $(a,b)\in R(F)$).

% AFs can be easily represented as a directed graph. 
The instantiation of an AF now consists of two parts, %reflecting the pair structure of the AF, 
namely the argument construction and the attack 
relation construction. An argument $A=(S,C)$ consists of a support for the argument and a claim. The support is a subset of the knowledge base $\knowledgebase$ 
and the claim is a single logical formula. %contained in $\presetclaims$. 
The support must be a consistent and subset minimal set of formulae, which entails the claim. 
% Lastly the claim 
% must be contained in a pre-defined set of claims $\presetclaims$.

% \begin{definition}
% \label{def:arg}
% An argument is a pair $A = (S,C)$ s.t.
% \begin{enumerate}
%  \item $S \subseteq \knowledgebase$ is a consistent set of formulae
%  \item $S \models C$
%  \item $\not \exists \; S' \subset S$ s.t. $S' \models C$
%  \item $C \in \presetclaims$
% \end{enumerate}
% We call $S$ the support and $C$ the claim of $A$.
% \end{definition}

Here the arguments are pairs of support and claim to provide a formal basis for argument construction. When plugged in the argumentation 
framework we abstract from this ``inner'' structure 
and collapse every pair of support and claim into one abstract argument. This is the abstraction procedure of the overall process.

We note that argument construction here differs from the usual argument definition in the literature. In particular the claim can be taken only from a pre-defined set $\presetclaims$. Using a 
pre-defined set of claims, we can restrict ourselves to reasonable claims, e.g. not involving tautologies. 
%In this way we can easily 
In this way we prohibit the construction of infinitely many arguments  
that could otherwise result from infinitely many syntactically different formulae which are semantically equivalent. 
%  which may arise otherwise. 
This restriction comes with a disadvantage however, as the set of pre-defined claims must be chosen with care, since inconsistent conclusions might be drawn otherwise. 
% undesired results may occur if certain arguments are not constructed. 
Indeed, \cite{GorogiannisH11} identify conditions for rational and consistent end results, which require the existence 
of specific arguments, which must be included in $\presetclaims$. %A pre-defined set of claims must now include all the necessary arguments. %, or otherwise inconsistent conclusions might be drawn. 
On the other hand, this restriction is in line with the concept of cores of argumentation frameworks \cite{AmgoudBV11}, which try to preserve 
desired properties while using only a subset of all possible arguments.

% The by the fourth condition. Without this condition we
% immediately deal with an infinite number of arguments with e.g. a tautologocial claim. In this work we simplify the process 
% by specifying a set of claims ($\presetclaims$) in advance, which may be derived. Note that the careful selection of $\presetclaims$ is important, since 
% inconsistent conclusions may be accepted in the end, depending on the attack relation construction. Indeed \cite{GorogiannisH11} identify conditions on the 
% attack relation for consistent end results. If we would restrict the set of claims to the elements in the knowledge base in Example~\ref{ex:kb} then we may 
% accept the whole knowledge base, which is obviously inconsistent and not desired. We note that the idea of restricting the set of claims is in line with 
% the concept of AF cores \cite{AmgoudBV11}, 
% which intuitively specify a set of arguments that are sufficient to ensure that rational properties, as consistency and closure, are fullfilled.
% \todo{JW: rewrite paragraph, see notes}

\begin{example}
\label{ex:arg_construction}
Continuing Example~\ref{ex:kb}, let the set of claims be $\presetclaims = \knowledgebase \cup \{\neg a, b, a \land \neg b\}$. Then we can construct the 
following arguments:
\small 
\begin{align*}
\qquad a_1 & = (\{a\},a) & \qquad a_4 & = (\{\neg b, a \rightarrow b\},\neg a) &\\
a_2 & = (\{a \rightarrow b\},a \rightarrow b) & a_5 & = (\{\neg b\}, \neg b) &\\
a_3 & = (\{a, a \rightarrow b\} ,b) & a_6 & = (\{a, \neg b\},a \land \neg b) &
\end{align*}
\normalsize 
% If we extend the claims by $\presetclaims' = \presetclaims \cup \{a \lor \neg b\}$ then we can additionally construct the following two arguments.
% \begin{eqnarray*}
%  a_7 & = & (\{a\},a \lor \neg b)\\
%  a_8 & = & (\{\neg b\},a \lor \neg b)
% \end{eqnarray*}
% Note that both $a_7$ and $a_8$ have the same claim, but are supported by different subsets of the knowledge base.
\end{example}

% \todo{simplification by pre-defined claims, rationality, core}

% \todo{example arguments from example 1}

For the construction of the attack relation several options were studied in literature. The basic idea for attacks between arguments underlying 
all of these options is that some sort of inconsistency occurs between them. We take the attack definitions from \cite{GorogiannisH11} and illustrate two 
types, defeat and directed defeat. An argument $A = (S,C)$ attacks an argument $A' = (S'=\{\phi_1',...,\phi_m'\},C')$ using defeat if $C \models \neg (\phi_1' \land ... \land \phi_m')$. The former directly defeats the latter 
if $C \models \neg \phi_i'$ for one $i$, $1 \leq i \leq m$. %\todo{JW: direct defeat instead of rebut}
% \todo{strong rebut from Amgoud Besnard, 2010?}

% \todo{simplify and take only 2/3 defs}
% \begin{definition}
% \label{def:attack}
% Let $A = (S,C)$ and $A' = (S'=\{\phi_1',...,\phi_m'\},C')$ be two arguments, then \todo{defeat rebuts name?}
% \begin{eqnarray*}
%  A \text{ rebuts } A' & \text{if} & C \equiv \neg C'\\
%  A \text{ defeat rebuts } A' & \text{if} & C \models \neg C'\\
%  A \text{ defeats } A' & \text{if} & C \models \neg (\phi_1' \land ... \land \phi_m')\\
%   A \text{ directly defeats } A' & \text{if} & C \models \neg  \phi_i' \text{ for a }\phi_i' \in S'\\
%   A \text{ undercuts } A' & \text{if} & C \equiv \neg (\phi_1'' \land ... \land \phi_t'') \text{ for } \{\phi_1'',...,\phi_t''\} \subseteq S'\\
%   A \text{ undercuts (variant) } A' & \text{if} & C \cup S' \models \bot\\
%   A \text{ directly undercuts } A' & \text{if} & C \equiv \neg \phi_i' \text{ for a } \phi_i' \in S'\\
%   A \text{ canonically undercuts } A' & \text{if} & C \equiv \neg (\phi_1' \land ... \land \phi_m')\\
%   A \text{ strongly rebuts } A' & \text{if} & S \cup S' \models \bot
% \end{eqnarray*}
% 
% \end{definition}

\begin{example}
\label{ex:af_constructed}	
% \todo{attack relation example continued, nicer example structure}
% \todo{g:not all relations mentioned here, e.g. a2 rebuts a5 and vice versa, just mention that those are examples for attacks}
% Continuing Example~\ref{ex:arg_construction}, then
% \begin{itemize}
%  \item $a_1$ (defeat) rebuts $a_4$ and vice versa
%  \item $a_4$ (directly) undercuts $a_1$, but not vice versa
%  \item $a_4$ and $a_5$ strongly rebut each other
% \end{itemize}
Continuing Example~\ref{ex:arg_construction}, the AF in Figure~\ref{fig:AF_undercut} illustrates the result using the direct defeat on the arguments built from $\knowledgebase$ and the claims $\presetclaims$. 
Note that e.g. $a_3$ and $a_5$ are not mutually attacking each other, since the claim of $a_5$ does not entail a negated support formula of $a_3$.
\vspace{-2em}
\begin{figure}
\begin{center}
\caption{Argumentation framework}
\label{fig:AF_undercut}
\begin{tikzpicture}[scale=1,>=stealth']
		\path 	node[arg](1){$a_1$}
			++(0,-1) node[arg,inner sep=2.8](4){$a_4$}
			++(-1,-1) node[arg](5){$a_3$}
			++(2,0) node[arg,inner sep=1.8](6){$a_6$}
			++(-2.75,-0.75) node[arg,inner sep=3](2){$a_5$}
			++(3.5,0) node[arg](3){$a_2$}
			;
		\path [left,->, thick]
			(4) edge (1)
			(4) edge (5)
			(4) edge (6)
			(5) edge (2)
			(5) edge (4)
			(5) edge (6)
			(6) edge (3)
			(6) edge (4)
			(6) edge (5)
			;
% 		\path [bend left, left, above,->, thick]
% 			(c) edge (d)
% 			(d) edge (c);
\end{tikzpicture}
\end{center}
\end{figure}
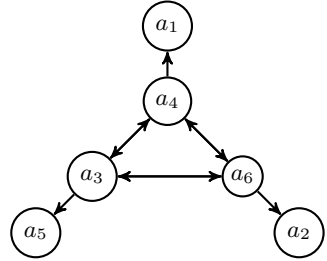
\vspace{-3em}
\end{example}

% \todo{attack constr... some defs}
% \todo{simplify semantics and only mention that they exist}
This completes the first step of the argumentation process, namely the AF construction out of the knowledge base. For the conflict resolution a plethora 
of argumentation framework semantics exist. A basic property for semantics is the conflict-free property, which states that a set $M$ of arguments in an AF $F$ is conflict free if 
there are no attacks between them in $F$. A set of arguments $M$ is stable in an AF $F = (\mathit{Args},\mathit{Att})$ if it is conflict free and all arguments outside are attacked from $M$, i.e. 
$\forall a \in (Args \setminus M) \;\exists b \in M$ with $(b,a) \in \mathit{Att}$.

\begin{example}
\label{ex:exts}
If we take the argumentation framework from Example~\ref{ex:af_constructed}, then the stable semantics selects $\big\{ \{a_1,a_5,a_6\} ,\{a_1,a_2,a_3\}, \{a_2,a_4,a_5\} \big\}$ as acceptable subsets of arguments.
% \begin{eqnarray*}
% \stable = \big\{ \{a_1,a_5,a_6\} ,\{a_1,a_2,a_3\}, \{a_2,a_4,a_5\} \big\}
% \end{eqnarray*}
\end{example}

% \begin{example}\label{example:semantics}
% Consider the AF $\AF$ from Example \ref{example:AF}. 
% %
% We have $\{a,d,f\}$ and $\{a,c,f\}$ as the stable extensions and thus %that 
% $\stable(F)=\stage(F)=\semi(F)=\{\{a,d,f\},\{a,c,f\}\}$. 
% The admissible sets of $F$ are
% $\{\}$, $\{a\}$, $\{c\}$, $\{a,c\}$, $\{a,d\}$, $\{c,f\}$, $\{a,c,f\}$, $\{a,d,f\}$
% and therefore  $\pref(F)=\{\{a,c,f\}$,$\{a,d,f\}\}$.
% %
% Finally we have  $\comp(F)=\{\{a\}$, $\{a,c,f\}$, $\{a,d,f\}\}$, with $\{a\}$ being the grounded extension.
% \end{example}

The last step of the argumentation process deals with drawing conclusions from the sets of acceptable arguments. One can look at the content of the abstract arguments 
which were accepted, e.g. one can derive the deductive closure of this content.
  
% \begin{example}
% From the extension from Example~\ref{ex:exts} $\{a_1,a_2,a_6\}$ we can conclude that an acceptable view of the knowledge base is to take $a,\neg b, a \land \neg b$ using 
% the stable semantics. %Taking a more skeptical approach the grounded semantics we only have the empty unique grounded extension from which we can 
% % conlude only tautologies using classical logic.
% \end{example}

In general every step of this process is intractable. Hence we need sophisticated systems for tackling these steps, which makes ASP a suitable choice for embedding the process in. 
A more detailed computational complexity analysis can be found in \cite{ParsonsWA03}.

% \todo{Complexity results of whole instantiation process. reference besnard, hunter woltran? check where this result is}

% what belongs here:
% short intro
% argumentation process
%  KB:            many possibilities, we focus here on PL0
%  instantiation: fundamental model is AF, still options:
%                 .) args -> very much fixed (s,c)
%                 simplification: assume set of claims given, this may disrupt some results!, but is ``inline'' of core-concept, which is required beforehand...
%                 .) att  -> many options
%                            describe variants
%  conclusion:    ``somewhat'' fixed for KB: (deductive closure?)

\subsection{Answer-Set Programming} \label{sec:asp}
In this section we recall the basics % the syntax and
%semantics 
of disjunctive logic programs under the answer-sets semantics
\cite{BrewkaET11,niem-99}. %Then we proceed to modeling techniques for so-called loops over sets and we briefly outline saturation. 
% Lastly we will recall complexity results relevant for our purposes.

We fix a countable set $\U$ of {\em (domain) elements}, also called \emph{constants}.
% and suppose a total order $<$ over the domain elements.
An {\em atom} is an expression
$p(t_{1},\ldots,t_{n})$, where $p$ is a {\em predicate} of arity $n\geq 0$
and each $t_{i}$ is either a variable or an element from $\U$.
An atom is \emph{ground} if it is free of variables.
$\BU$ denotes the set of all ground atoms over $\U$.
%A {\em (classical) literal} $l$
%is an atom $a$ (in this case, it is {\em positive}), or a
%negated atom $\tneg a$ (in this case, it is {\em negative}).
%Given a literal $l$, its {\em complement} $\tneg l$
%is defined as $\tneg a$ if $l=a$ and $a$ if $l=\tneg a$.
%A set $L$ of literals is said to be {\em consistent} if, for every
%literal $l \in L$, $\tneg l \notin L$.

A \emph{(disjunctive) rule} $r$ is of the form
% \begin{center}
% \vspace{-2pt}
\begin{equation*}
a_1\ \vee\ \cdots\ \vee\ a_n\ \la
        b_1,\ldots, b_k,\
        \naf b_{k+1},\ldots,\ \naf b_m
\end{equation*}
% \end{center}
% \vspace{-2pt}
with $n\geq 0,$ $m\geq k\geq 0$, $n+m > 0$, where
$a_1,\ldots ,a_n,b_1,\ldots ,b_m$ are
atoms, and ``$\naf$'' stands for {\em default negation}.
%literals.
%We refer to ``$\neg$'' as {\em strong negation} and to ``$\naf$'' as
%{\em default negation}, or simply as {\em negation}.
The \emph{head} of $r$ is the set
$\head{r}$ = $\{a_1\commadots a_n\}$ and
the \emph{body} of $r$ is
$\body{r}=
\{b_1,\ldots, b_k,$ $\naf b_{k+1},\ldots,$ $\naf  b_m\}$.
Furthermore, $\bodyp{r}$ = $\{b_{1}\commadots b_{k}\}$ and
$\bodyn{r}$ = $\{b_{k+1}\commadots b_m\}$.
A rule $r$ is a %\emph{normal} if $n \leq 1$ and a %(hard) 
\emph{constraint} if $n=0$. 
A rule $r$ is \emph{safe} if each variable in $r$ occurs in $\bodyp{r}$.
A rule $r$ is \emph{ground} if no variable occurs in $r$.
A \emph{fact} is a ground rule without disjunction and empty body. 
% An \emph{(input) database} is a set of facts.
A program is a finite set of disjunctive rules. 
% For a program $\p$ and an input database $D$, we often write $\p{(D)}$ instead of $D\cup\p$.
% If each rule in a program is 
% normal (resp.\ ground), we call the program normal (resp.\ ground).
% A program $\p$ is called \emph{stratified}  if there exists an assignment $a(\cdot)$
% of integers to the predicates in $\p$, such that for each $r\in\p$, the following holds:
% If predicate $p$ occurs in the head of $r$ and predicate $q$ occurs 
% (i) in the positive body of $r$, then $a(p)\geq a(q)$ holds;
% (ii) in the negative body of $r$, then $a(p)> a(q)$ holds.

For any program \p{}, let \UP{}
be the set of all constants appearing in \p{}.
%(if no constant appears in \p, an arbitrary constant
%is added to \UP)
%let \BP{} be the set of all ground
%literals constructible from the predicate
%symbols appearing in \p{} and the constants of \UP{}.
%Moreover,
$\GP$ is  the set of rules
$r\sigma$
obtained by applying, to each rule 
%and weak constraint
$r\in\p{}$, all possible
substitutions $\sigma$ from the variables
in $r$ to elements of $\UP{}$.
%We call $\Ground{\p,U_P}$ the grounding of $\p{}$, and
%write $\GP$ as a shorthand for $\Ground{\p,U_P}$.
%$\UP{}$ is usually called the \emph{Herbrand Universe} of \p{} and
%$\BP{}$
%the \emph{Herbrand Literal Base} of \p{}.
%
An \emph{interpretation} $I\subseteq \BU$ 
\emph{satisfies} %by %a consistent set of literals 
%an interpretation $I$
a ground rule $r$
iff $\head{r} \cap I \neq \emptyset$ whenever
$\bodyp{r}\subseteq I$ and $\bodyn{r} \cap I = \emptyset$.
$I$ satisfies a ground program $\p$,
if each $r\in\p$
is satisfied by $I$.
%A ground (weak) constraint $c$ is {\em violated} by $I$,
%iff $\bodyp{c}\subseteq I$ and $\bodyn{c}\cap I=\emptyset$. 
%%it is satisfied otherwise.
A non-ground rule $r$ (resp., a program $\p$)
is satisfied by an interpretation $I$ iff
$I$ satisfies all groundings of $r$ (resp., $\GP$).
%A non-ground (weak) constraint
%$c$ is violated by
%$I$ iff
%$I$ violates at least one grounding of $c$.
%
%For programs $\p$ without weak constraints, 
$I \subseteq \BU$ is an \emph{answer-set}
%\footnote{\label{fn}Note that we only consider {\em consistent
%answer sets}, while in \cite{gelf-lifs-91} also the inconsistent set of all
%possible literals can be a valid answer set.}
of $\p$
iff it is a subset-minimal set
satisfying
the \emph{Gelfond-Lifschitz reduct}
$
\p^I=\{ \head{r} \derives \bodyp{r} \mid I\cap
\bodyn{r} = \emptyset, r \in \GP\}
$.

\section{Instantiation-based Argumentation}
\label{sec:inst_based_argu}

In this section we provide our ASP encodings for the construction of arguments from a knowledge-base $\knowledgebase$ and a set $\presetclaims$ of claims. As input, each formula in $\knowledgebase$ and $\presetclaims$ is given by the unary predicate $\kb(\cdot)$ and $\cl(\cdot)$, respectively. 
\begin{example} \label{ex:aspinput}
The input, as given in Example \ref{ex:kb} and \ref{ex:arg_construction}, is specified by: 
\begin{center}
\small
$\{\kb(a).$ $\kb(\fimp(a,b)).$ $ \kb(\fneg(b)). $ \\
$\cl(a).$ $ \cl(\fimp(a,b)).$ $ \cl(\fneg(b)).$ $ \cl(\fneg(a)).$ $ \cl(b).$ $ \cl(\fand(a,\fneg(b))). \}$
\end{center}
\end{example}

First, we introduce the ASP encodings for checking whether a certain variable assignment is a model for a given formula (or not). Model checking plays a crucial role for our instantiation-based approach. Then, we present encodings for the computation of arguments. Finally, we provide ASP code for some types of attack relations. 
Note that an argumentation framework is obtained by two separate ASP program calls where the first one takes as input $\knowledgebase$ and $\presetclaims$ and returns a separate answer-set for each resulting argument. The second program receives as input a ``flattened'' version of all arguments and computes the attacks between arguments based on different attack type encodings.

\subsection{Model Checking}
Propositional formulae provide the basis for the construction of arguments and their attack relations. In fact, we can express most of the defining properties of arguments (such as entailment of the support to the claim) and attacks by means of propositional formulae. In this section we provide an ASP encoding that allows us to check whether a formula $\alpha$ is true under a given interpretation $I$, i.e. $I$ is a model for $\alpha$. First, the formula is split into sub-formulae until we obtain the contained atoms or constants. Due to brevity, the following encodings  only exemplify this for the connectives $\land$, $\neg$ and $\rightarrow$. Note that $\lor$, $\not \leftrightarrow$ and $\leftrightarrow$ are supported as well.
%\todo{source for model checking}
\begin{asp}{\subformula}{asp:subformula}
\subformula(F) \la \subformula(\fand(F,\_));\\
&& \subformula(F) \la \subformula(\fand(\_,F));\\
%&& \subformula(F) \la \subformula(\for(\_,F));\\
%&& \subformula(F) \la \subformula(\for(F,\_));\\
&& \subformula(F) \la \subformula(\fneg(F));\\
%&& \subformula(F) \la \subformula(\fxor(F,\_));\\
%&& \subformula(F) \la \subformula(\fxor(\_,F)).
&& \subformula(F) \la \subformula(\fimp(F,\_));\\
&& \subformula(F) \la \subformula(\fimp(\_,F)).
%&& \subformula(F) \la \subformula(\fiff(F,\_));\\
%&& \subformula(F) \la \subformula(\fiff(\_,F)).
\end{asp}

\noindent The atoms and constants of $\alpha$ are then obtained via the encoding $\pi_\atom$. Consider 
rule (1) which denotes that a formula is not an atom in case it is of the form $\fand(\cdot,\cdot)$
\footnote{Note that the syntax of our encodings is specific to the grounder gringo \cite{GebserKKOTM11}.
}.
\begin{asp}{\atom}{asp:atom}
\noatom(F) \la \subformula(F;F1;F2), F:=\fand(F1,F2);\\
%&& \noatom(F) \la \subformula(F;F1;F2), F:=\for(F1,F2);\\
&& \noatom(F) \la \subformula(F;F1), F:=\fneg(F1);\\
%&& \noatom(F) \la \subformula(F;F1;F2), F:=\fxor(F1,F2);\\
&& \noatom(F) \la \subformula(F;F1;F2), F:=\fimp(F1,F2);\\
%&& \noatom(F) \la \subformula(F;F1;F2), F:=\fiff(F1,F2);\\
&& \atom(X) \la \subformula(X), \naf \noatom(X).
%&& \atomc(X) \la \subformula(X), X:=\fc(\top);\\
%&& \atomc(X) \la \subformula(X), X:=\fc(\bot).
\end{asp}

Now we compute whether the interpretation is a model by first evaluating the atoms and constants. In case an atom gets assigned $\ptrue$ ($\pfalse$) we derive that the interpretation for this sub-formula is a model (not a model). Now, the connectives are evaluated bottom-up based on the model information of the sub-formulae. In particular, this allows to check whether $I$ is a model for our original formula $\alpha$, or not.

The encoding $\pi_\ismodel$ exemplifies this approach for some of the connectives. 
In the subsequent sections we have to apply model checking several times within a single ASP encoding. In order to avoid side effects of different checks, we introduce an additional parameter, $K$, which serves as a key for identifying the origin of the interpretation that is currently checked. Suppose, for example, that we want to check satisfiability of two different formulae. As the formulae may evaluate to true under different interpretations we have to distinguish between the truth assignments.

\begin{aspworkaround}{asp:ismodel}
&&\pi_{\ismodel}  = \big \{  \ismodel(K,X) \la \atom(X), \ptrue(K,X);\\
%&&\hphantom{\pi_{\ismodel} = \big \{} \ismodel(K,X) \la \atomc(X), X:=\fc(\top), \key(K);\\
&& \! \begin{split}\hphantom{\pi_{\ismodel} = \big \{} & \ismodel(K,F) \la \hspace{-1em} && \subformula(F;F1), F:=\fneg(F1), \\
&&& \nomodel(K,F1);\end{split}\\
&& \! \begin{split}\hphantom{\pi_{\ismodel} = \big \{} & \ismodel(K,F) \la \hspace{-1em} && \subformula(F), F:=\fand(F1,F2), \\
&&& \ismodel(K,F1;F2);\end{split}\\
%&& \ismodel(K,F) \la \subformula(F;F1;F2), F:=\for(F1,F2), \ismodel(K,F1);\\
%&& \ismodel(K,F) \la \subformula(F;F1;F2), F:=\for(F1,F2), \ismodel(K,F2);\\
%&& \ismodel(K,F) \la \subformula(F), F:=\fxor(F1,F2),\\
%&& \hphantom{\ismodel(K,F) \la} \ismodel(K,F1), \nomodel(K,F2);\\
%&& \ismodel(K,F) \la \subformula(F), F:=\fxor(F1,F2), \\
%&& \hphantom{\ismodel(K,F) \la} \ismodel(K,F2), \nomodel(K,F1).
%&& \ismodel(K,F) \la \subformula(F;F1;F2), F:=\fimp(F1,F2), \nomodel(K,F1);\\
&& \! \begin{split}\hphantom{\pi_{\ismodel} = \big \{} & \ismodel(K,F) \la \hspace{-1em} && \subformula(F), F:=\fimp(F1,F2), \\
&&& \ismodel(K,F1;F2). \; \big \} \end{split}
%&& \ismodel(K,F) \la \subformula(F), F:=\fiff(F1,F2), \ismodel(K,F1;F2);\\
%&& \ismodel(K,F) \la \subformula(F), F:=\fiff(F1,F2), \nomodel(K,F1;F2).
\end{aspworkaround}

% nomodel for space reasons commented out
% \begin{asp}{\nomodel}{asp:nomodel}
% \nomodel(K,X) \la \atom(X), \pfalse(K,X);\\
% %&& \nomodel(K,X) \la \atomc(X), X:=\fc(\bot), \key(K);\\
% && \nomodel(K,F) \la \subformula(F;F1), F:=\fneg(F1), \\
% && \hphantom{\nomodel(K,F) \la} \ismodel(K,F1);\\
% && \nomodel(K,F) \la \subformula(F;F1;F2), F:=\fand(F1,F2), \\
% && \hphantom{\nomodel(K,F) \la} \nomodel(K,F1);\\
% && \nomodel(K,F) \la \subformula(F;F1;F2), F:=\fand(F1,F2), \\
% && \hphantom{\nomodel(K,F) \la} \nomodel(K,F2);\\
% %&& \nomodel(K,F) \la \subformula(F), F:=\for(F1,F2), \nomodel(K,F1;F2);\\
% %&& \nomodel(K,F) \la \subformula(F), F:=\fxor(F1,F2), \\
% %&& \hphantom{\nomodel(K,F) \la} \ismodel(K,F1;F2);\\
% %&& \nomodel(K,F) \la \subformula(F), F:=\fxor(F1,F2), \\
% %&& \hphantom{\nomodel(K,F) \la} \nomodel(K,F1;F2).
% && \nomodel(K,F) \la \subformula(F), F:=\fimp(F1,F2), \\
% && \hphantom{\nomodel(K,F) \la}\ismodel(K,F1), \nomodel(K,F2).
% %&& \nomodel(K,F) \la \subformula(F), F:=\fiff(F1,F2), \nomodel(K,F1), \ismodel(K,F2);\\
% %&& \nomodel(K,F) \la \subformula(F), F:=\fiff(F1,F2), \nomodel(K,F2), \ismodel(K,F1).
% \end{asp}

Due to brevity we omit the encoding $\pi_\nomodel$ here. Analogous to $\pi_\ismodel$ it derives the predicate $\nomodel(K,F)$ whenever an atom gets assigned $\pfalse$ or a sub-formula is false under the current interpretation. 
The complete program for checking whether a formula evaluates to true under a given variable assignment consists of 
\begin{center}
\small 
$\pi_{\modelcheck} = \pi_{\subformula} \cup \pi_{\atom} \cup \pi_{\ismodel} \cup \pi_{\nomodel}$
\end{center} 

\begin{example}
Consider the formula $a \rightarrow b$ from $\knowledgebase$ of Example~\ref{ex:aspinput}, \ie $\kb(\fimp(a,b))$. In order to check whether there exists a model we can make use of $\pi_{\modelcheck}$ in the following way: Initially, we have to define an additional rule $\subformula(X) \la \kb(X)$ as $ \pi_{\subformula}$ only considers formulae given by the predicate $\subformula(\cdot)$. By adding the program $\pi_{\subformula} \cup \pi_{\atom}$ the following answer-set is returned:
\begin{center}
\small
$\{\kb(\fimp(a,b)).$ $\subformula(\fimp(a,b)).$ $\subformula(a).$ $\subformula(b). $ \\
$\noatom(\fimp(a,b)).$ $\atom(b).$ $\atom(a).\}$
\end{center}
\noindent Each atom now gets assigned true or false, representing an interpretation. We encode this by the rule $\ptrue(k, X) \lor \pfalse(k, X) \la \atom(X).$ Note that the specification of a key (in this case $k$) is mandatory although $\pi_{\modelcheck}$ is not applied several times in this example. By adding and running $\pi_{\ismodel} \cup \pi_{\nomodel}$ four answer-sets are returned. Each contains the predicates from the previously given answer-set as well as the truth assignment for the atoms $a$ and $b$ and either $\ismodel(k,\fimp(a,b))$ or $\nomodel(k,\fimp(a,b))$. The answer-set obtained by $\pfalse(k, a)$ and $\ptrue(k, b)$ contains (amongst others)
\begin{center}
\small
$\{\pfalse(k,a).$ $\ptrue(k,b).$ $\ismodel(k,b).$ $\nomodel(k,a).$ $\ismodel(k,\fimp(a,b)). \}$
\end{center}
denoting that $I(a) = \pfalse, I(b) = \ptrue$ is a model for $a \rightarrow b$.
\end{example}

\subsection{Forming Arguments}

We now derive the arguments from a knowledge base $\knowledgebase$ and a set $\presetclaims$ of claims. According to \cite{BesnardH01}, we have to check whether the support entails the claim and if the support is subset minimal as well as consistent.
%As stated in Section~\ref{sec:argumentation} we derive arguments from a knowledge base $\knowledgebase$ and a set $\presetclaims$ of claims. As input, each formula in $\knowledgebase$ and $\presetclaims$ is given by the unary predicate $\kb(\cdot)$ and $\cl(\cdot)$ respectively. 
In order to obtain arguments we first guess exactly one claim and a subset of formulae from $\knowledgebase$. This \emph{guess} is encoded as follows:
\begin{asp}{\arg}{asp:guess_argument}
 1 \{\; \pclaim(X) : \cl(X) \; \} 1;\\
&& 1 \{ \; \fs(X) : \kb(X) \; \}. %\}
\end{asp}
The selected claim is denoted by $\pclaim(\cdot)$.
The predicate $\fs(\cdot)$ is derived if the respective formula from $\knowledgebase$ is contained in the support $S$ of an argument $A = (S,C)$.

\paragraph{Entailment:} In order to be a valid argument, the support must entail the claim, \ie $S \models C$ must hold. 
As $S \models C$, $\models S \rightarrow C$ must hold as well. Hence, $\neg (S \rightarrow C) \equiv \neg (\neg S \lor C) \equiv S \land \neg C$ must be unsatisfiable. Unsatisfiability 
of the formula $S \land \neg C$ 
can be checked by making use of the saturation technique~\cite{EiterG95}:
 We first assign $\ptrue(\kentail, x)$ or $\pfalse(\kentail, x)$ to each atom $x$ in the formula using a disjunctive rule. This allows \emph{both} $\ptrue(\kentail, x)$ and $\pfalse(\kentail, x)$ to be contained in the resulting answer-set. Furthermore, all formulae in $S$ and the negated claim $C$ are conjunctively connected. Hence, in case any of those formulae evaluates to false under a variable assignment (\ie $\nomodel(\kentail, \cdot )$ is derived) we know that $\neg (S \rightarrow C)$ is not satisfied which implies that $S \models C$ evaluates to true under the given interpretation. In this case we \emph{saturate}, \ie we derive $\ptrue(\kentail, x)$ and $\pfalse(\kentail, x)$ for any atom $x$. On the other hand, if no formula in $S$ and $C$ derives $\entailsclaim$ the constraint $\la \naf \entailsclaim$ removes the answer-set. If this is the case, due to the definition of stable model semantics in answer-set programming, no answer-set is returned. Only in case there exists no model for  $\neg (S \rightarrow C)$ all guesses are saturated and we obtain a single answer-set representing a support $S$ and claim $C$ where $S \models C$ holds.

In the following the program $\pi_{\entailment}$ is given. Note that $\kentail$ is simply used as a key for identifying the variable assignment and model check.

\begin{asp}{\entailment}{asp:entail}
\ptrue(\kentail, X) \lor \pfalse(\kentail, X)  \la \atom(X);\\
% check if the guess is not a model for each fs and the negated claim
&& \entailsclaim \la \nomodel(\kentail, neg(X)), \pclaim(X);\\
&& \entailsclaim \la \nomodel(\kentail, X), \fs(X);\\
% constraint if fs's do not entail claim
&& \la \naf \entailsclaim;\\
% saturate (this ensures that we consider every interpretation!)
&& \ptrue(\kentail, X) \la \entailsclaim, \atom(X);\\
&& \pfalse(\kentail, X) \la \entailsclaim, \atom(X).
\end{asp}

\paragraph{Subset minimality:}The support $S$ of an argument must be a subset minimal set of formulae, \ie there must not exist an $S' \subset S$ s.t. $S' \models C$. 
Here, we apply the concept of a \emph{loop} (see e.g. \cite{EiterIK09}). 
For a candidate support $S$ we consider all $S' \subset S$ where there exists exactly one formula $\alpha \in S$ but $\alpha \not \in S'$. 
In case any such $S'$ exists where $S' \models C$ we know that $S$ is not a support for $C$. 
Due to monotonicity of classical logic this is sufficient since if $S' \not \models C$ then also for all $S'' \subset S'$ it holds that $S'' \not \models C$.
First, we define a total ordering over all formulae $\fs(\cdot)$ in $S$:
\begin{asp}{\order}{asp:order}
% \label{asp:order}
% \begin{eqnarray*}
% \pi_\order & =   \{ &
       \lt(X,Y)        \la \fs(X),\fs(Y),X<Y; \\
&&      \nsucc(X,Z)     \la \lt(X,Y),\lt(Y,Z); \\
&&      \succ(X,Y)      \la \lt(X,Y),\naf \nsucc(X,Y); \\
&&      \ninf(Y)        \la \lt(X,Y);\\
&&      \inf(X)         \la \fs(X), \naf \ninf(X);\\
&&      \nsup(X)        \la \lt(X,Y); \\
&&      \sup(X)\la      \fs(X), \naf \nsup(X).
% \end{eqnarray*}
\end{asp}

For any $S'$ we now assign $\ptrue(\kmin(K),x)$ or $\pfalse(\kmin(K),x)$ to all atoms $x$.  $\kmin(K)$ is used as key for identifying the truth assignment. $K$ is the formula $\alpha$ where $\alpha \not \in S'$. The idea is now to ``iterate'' over the the ordering, beginning at the infimum $\inf(\cdot)$. Based on the ordering, we now consider every formula from the support: In case the formula is satisfied or corresponds to the removed formula $\alpha$ (\ie the key $K$) we derive $\modelupto(\kmin(\alpha),\cdot)$. If we can derive $\hasmodel(\kmin(\alpha))$ we know that the support $S' = S \setminus \alpha$ is satisfiable and can therefore not be a valid support for our claim. On the other hand, if any $S'$ is a valid support we can not derive $\hasmodel(\kmin(\alpha))$ and the answer-set is removed by the constraint $\la \naf \hasmodel(\kmin(\alpha)), \fs(\alpha)$. %The program $\pi_{\mini} \cup \pi_{<}$ checks the subset minimality of a candidate support.

\begin{aspworkaround}{asp:minimize}
&& \pi_{\mini}  = \big \{ \ptrue(\kmin(K),X) \la \naf \pfalse (\kmin(K),X), \atom(X), \fs(K);\\
&& \hphantom{\pi_{\mini} = \big \{} \pfalse (\kmin(K),X) \la \naf \ptrue (\kmin(K),X), \atom(X), \fs(K);\\
% check from inf to sup if the guess in,out is a model, we do not check if model_upto(X,X)
&& \! \begin{split}\hphantom{\pi_{\mini} = \big \{} & \modelupto(\kmin(K),X) \la \hspace{-1em} && \inf(X), \ismodel(\kmin(K),X), \\
&&& \fs(X), X \not = K;\end{split}\\
&& \hphantom{\pi_{\mini} = \big \{} \modelupto(\kmin(K),K) \la \inf(K), \fs(K);\\
&& \! \begin{split}\hphantom{\pi_{\mini} = \big \{} & \modelupto(\kmin(K),X) \la \hspace{-1em} && \succ(Z,X), \ismodel(\kmin(K),X), \\
&&& \fs(X), \modelupto(\kmin(K),Z), X \not = K;\end{split}\\
&& \! \begin{split}\hphantom{\pi_{\mini} = \big \{} & \modelupto(\kmin(K),K) \la \hspace{-1em} && \succ(Z,K), \modelupto(\kmin(K),Z), \\
&&& \fs(K);\end{split}\\
&& \! \begin{split}\hphantom{\pi_{\mini} = \big \{} & \hasmodel(\kmin(K)) \la \hspace{-1em} && \sup(K), \modelupto(\kmin(K),X), \\
&&& \ismodel(\kmin(K),\fneg(Z)), \pclaim(Z);\end{split}\\
&& \hphantom{\pi_{\mini} = \big \{} \la \naf \hasmodel(\kmin(K)), \fs(K). \; \big \}
\end{aspworkaround}

\paragraph{Consistency:} The support $S$ must be a consistent set of formulae. In other words, there exists a model for the conjunction of all formulae in $S$. The program $\pi_{\consistent}$ simply consists of a guess which assigns truth values to all atoms and a constraint that removes any unsatisfiable support.

\begin{asp}{\mathrm{consistent}}{asp:consistency}
% Consistency
1 \{ \; \ptrue(\consistent,X) , \pfalse(\consistent,X) \; \} 1 \la \atom(X).\\
&& \la \nomodel(\consistent,X), \fs(X). \; \}.
\end{asp}

The following program then gives all arguments that can be computed from a knowledge base $\knowledgebase$ and a set of claims $\presetclaims$:
\begin{center}
\small
$\pi_{\arguments} = \pi_{\modelcheck} \cup \pi_{\arg} \cup \pi_{\entailment} \cup \pi_{<} \cup \pi_{\mini} \cup \pi_{\consistent}$
\end{center}
Each answer-set obtained by $\pi_{\arguments}$ contains the predicate $\pclaim(\cdot)$ and a set of predicates $\fs(\cdot)$, representing claim and support.

\begin{example} \label{ex:attout}
Consider the input as given in Example~\ref{ex:aspinput}. The program $\pi_{\arguments} $ returns the following answer-sets (we restrict ourselves to the relevant predicates):
\small
\begin{eqnarray*}
 a_1: && \{\fs(a).\ \pclaim(a). \}\\
 a_2^*: && \{\fs(\fimp(a,b)).\ \pclaim(\fimp(a,b)).\} \\
 a_3: && \{\fs(a).\ \fs(\fimp(a,b)).\ \pclaim(b). \}\\
 a_4: && \{\fs(\fneg(b)).\ \fs(\fimp(a,b)).\ \pclaim(\fneg(a)). \}\\
 a_5: && \{\fs(\fneg(b)).\ \pclaim(\fneg(b)). \}\\
 a_6: && \{\fs(a).\ \fs(\fneg(b)).\ \pclaim(and(a,\fneg(b))). \}\\
\end{eqnarray*}
\normalsize
Note that due to the definition of program $\pi_{\mini}$ and $\pi_{\consistent}$ several  resulting answer-sets may represent the same derived argument: This is the case for $a_2^*$ where actually three models are derived by the program $\pi_{\consistent}$. They only differ in the respective truth assignments $\ptrue(\consistent,\cdot)$ and $\pfalse(\consistent,\cdot)$. We eliminate duplicates in an additional post-processing step in order to remove redundant information.
\end{example}

%%%%%%%%%%%%%%%%%%%%%%%%%%%

\subsection{Identifying Conflicts between Arguments}

We now want to compute attacks between arguments. Therefore we first specify  encodings that are used by every attack type (such as defeat and direct defeat). We then present encodings for the computation of these attack types. 

In order to reason over all arguments we first have to ``flatten'' the answer-sets obtained by $\pi_{\arguments}$. We specify this by the predicates $\as(A, fs, \cdot)$ and $\as(A, claim, \cdot)$. $A$ is a numeric key identifying the argument. 

\begin{example} \label{ex:flatten}
We illustrate this by the answer-sets $a_1$, $a_2$ and $a_3$ from Example~\ref{ex:attout}. This input is given by the following facts:
\begin{center}
\small 
$\{\as(1, \fs, a).$ $\as(1, \pclaim, a).$ $\as(2, \fs, \fimp(a,b)).$ $\as(2, \pclaim, \fimp(a,b)).$ $\as(3, \fs, a).$ $\as(3, \fs, \fimp(a,b)).$ $\as(3, \pclaim, b). \}$
\end{center}
\end{example}

%For each argument there may exist several predicates $\as(A, fs, \cdot)$ where the last parameter is a formula contained in the support of the argument. The last parameter of $\as(A, claim, \cdot)$ is the claim of the respective argument $X$. 

In order to identify conflicts between arguments we first guess two arguments. $\selectedA(\cdot)$ and $\selectedB(\cdot)$ contain the keys of the selected arguments. 

\begin{asp}{\mathrm{att}}{asp:guess_attack}
1 \{ \selectedA(A) : \as(A,\_,\_) \} 1.\\
&& 1 \{ \selectedB(A) : \as(A,\_,\_) \} 1. 
\end{asp}

Furthermore, we construct one single support formula for each argument $A$ by conjunction of all formulae in $\as(A, fs, \cdot)$. As in the previous section we first define an ordering over all formulae that are contained in the support. The only difference is that we add the argument's key $A$ to the predicates $\inf(A,\cdot)$, $\sup(A,\cdot)$ and $\succ(A,\cdot,\cdot)$. Due to brevity, the corresponding program $\pi_{<_{\key}}$ is omitted. We can then construct the support formula by iterating over the ordering and connecting the formulae by conjunction. Note that the last parameter of $\fsconj(A,\cdot,\cdot)$ is simply used as an identifier for the current position in the iteration. When the supremum is reached we derive $\support(A,\cdot)$ for $A$ containing the support formula.

\begin{asp}{\support}{asp:build_supp_form}
% if we would disallow self attacks:
%:- selected2(X), selected1(X).
% Order
% lt(A,X,Y) :- as(A,fs,X), as(A,fs,Y), X<Y.\\
% nsucc(A,X,Z) :- lt(A,X,Y), lt(A,Y,Z).\\
% succ(A,X,Y) :- lt(A,X,Y), not nsucc(A,X,Y).\\
% ninf(A,X) :- lt(A,Y,X).\\
% nsup(A,X) :- lt(A,X,Y).\\
% inf(A,X) :- not ninf(A,X), as(A,fs,X).\\
% sup(A,X) :- not nsup(A,X), as(A,fs,X).\\
\fsconj(A,X,X) \la \inf(A,X), \sup(A,X);\\
&& \fsconj(A,\fand(X,Y),Y) \la \inf(A,X), \succ(A,X,Y);\\
&& \fsconj(A,\fand(O,N),N) \la \succ(A,C,N), \fsconj(A,O,C);\\
&& \support(A,X) \la \fsconj(A,X,C), \sup(A,C). 
\end{asp}

For the computation of attacks we again apply the saturation technique. The program $\pi_{\attsat}$ is used to saturate all attack computations. First, we derive all attack type keys $t$ from the truth assignments $\ptrue(t,\cdot)$ and $\pfalse(t,\cdot)$ of the applied attack type programs. Note that the corresponding assignments are defined separately in each attack program. In case $\attack$ is derived for all truth assignments in some attack program we saturate. Finally, the binary predicate $\attack(\cdot, \cdot)$ is generated. It is used for the representation of the attack relation between the two arguments.

\begin{asp}{\attsat}{asp:attsat}
% Get all possible saturation keys for the used attack types
\attacktype(T) \la \ptrue(T,\_);\\
&& \attacktype(T) \la \pfalse(T,\_);\\
% saturate in case we have an attack
&& \ptrue(T,X) \la \attack, \atom(X), \attacktype(T);\\
&& \pfalse(T,X) \la \attack, \atom(X), \attacktype(T);\\
% in case we have no attack
&& \la \naf \attack;\\
% finally, generate attack between the two arguments for visualization
&& \attack(X,Y) \la \selectedA(X), \selectedB(Y).
\end{asp}

We now consider the attack types \emph{defeat} and \emph{direct defeat}. The basic idea is to define a propositional formula that represents the attack condition. We then assign $\ptrue(t,x)$ or $\pfalse(t,x)$ to any atom $x$ using a disjunctive rule. In case every such interpretation is a model for our attack formula we know that the formula is valid. Otherwise, if any interpretation is not a model (\ie $\attack$ is not derived) the resulting answer-set is strictly smaller than those where the interpretation is a model. Such answer-sets are removed by the constraint $ \la \naf \attack$. Note that this also works in case we consider several attack types: If we derive $\attack$ for all interpretations of a single attack type we saturate all interpretations of \emph{all} attack types. As the predicate $\attack$ as well as all assignments $\ptrue(t,x)$ and $\pfalse(t,x)$ are then contained in all answer-sets for the two selected arguments we derive every attack relation.
%\todo{ 	- implement  undercut
%	- implement undercut (variant)
%	- test strong rebuttal}

%The following program $\pi_{\rebuttal}$ exemplifies the above described approach. For two arguments $A = (S,C)$ and $A' = (S'=\{\phi_1',...,\phi_m'\},C')$ we have that $A$ rebuts $A'$ if $C \equiv \neg C'$. This is reflected by deriving $\checkRebuttal(\fiff(C,\fneg(C')))$. $\subformula(X) \la \checkRebuttal(X)$ simply guarantees that we can apply $\pi_{\modelcheck}$ to our formula.
%
%\begin{asp}{rebut}{asp:rebut}
%\checkRebuttal(\fiff(C,\fneg(C'))) \la \selectedA(X),  \selectedB(Y),\\
%&& \hphantom{\checkRebuttal(placeholder) }  \as(X,\pclaim,C), \as(Y,\pclaim,C');\\
%&& \subformula(X) \la \checkRebuttal(X);\\
%&& \ptrue(\rebuttal,X) \lor \pfalse(\rebuttal,X) \la \atom(X);\\
%&& \attack \la \ismodel(\rebuttal,X), \checkRebuttal(X).
%\end{asp}

%\begin{asp}{strong\_rebut}{asp:strongrebut}
%\checkStrongrebuttal(\fimp(\fand(A,B),\fc(\bot))) \la \selectedB(X), \\
%&& \hphantom{placeholder1 } \selectedA(Y), \support(X,A), \support(Y,B);\\
%&& \subformula(X) \la \checkStrongrebuttal(X);\\
%&& \ptrue(\strongrebuttal,X) \lor \pfalse(\strongrebuttal,X) \la \atom(X);\\
%&& \attack \la \ismodel(\strongrebuttal,X), \checkStrongrebuttal(X).
%\end{asp}

The following program $\pi_{\defeat}$ exemplifies the above described approach. For two arguments $A = (S,C)$ and $A' = (S'=\{\phi_1',...,\phi_m'\},C')$ we have that $A$ defeats $A'$ if $C \models \neg (\phi_1' \land ... \land \phi_m')$.
Here we make use of the support derived by $\pi_{<_\key} \cup \pi_{\support}$. As the support is defined as a conjunction of formulae we can directly make use of claim $C$ and the negated support $\neg S'$.
% \begin{asp}{\mathrm{defeat}}{asp:defeat}
% \begin{split}\checkDefeat(\fimp(C,\fneg(S'))) \la \selectedA(X),  \selectedB(Y),  \\
% \as(X,\pclaim,C), \support(Y,S');\end{split}\\
% && \subformula(X) \la \checkDefeat(X);\\
% && \ptrue(\defeat,X) \lor \pfalse(\defeat,X) \la \atom(X);\\
% && \attack \la \ismodel(\defeat,X), \checkDefeat(X).
% \end{asp}

\begin{aspworkaround}{asp:defeat}
&&\begin{split}\pi_{\mathrm{defeat}} & = \big \{ \checkDefeat(\fimp(C,\fneg(S'))) \la \hspace{-1em} && \selectedA(X),  \selectedB(Y),  \\
&&&\as(X,\pclaim,C), \support(Y,S');\end{split}\\
&&\hphantom{\pi_{\mathrm{defeat}} = \big \{} \subformula(X) \la \checkDefeat(X);\\
&&\hphantom{\pi_{\mathrm{defeat}} = \big \{} \ptrue(\defeat,X) \lor \pfalse(\defeat,X) \la \atom(X);\\
&&\hphantom{\pi_{\mathrm{defeat}} = \big \{} \attack \la \ismodel(\defeat,X), \checkDefeat(X). \; \big \}
\end{aspworkaround}

The second program we consider here is $\pi_{\directdefeat}$. $A$ directly defeats $A'$ if $C \models \neg  \phi_i'$ for a $\phi_i' \in S'$. Hence, we have to consider each formula in $S'$ separately. Therefore we use a combination of attack type and $\phi_i'$ to identify the truth assignment.
  
\begin{aspworkaround}{asp:directdefeat}
&& \begin{split}\pi_{\directdefeat} & = \big \{ \checkDirectdefeat(\Phi,\fimp(C,\fneg(\Phi))) \la \hspace{-1em} && \selectedA(X), \selectedB(Y),  \\
&&& \as(X,\pclaim,C), \as(Y,\fs,\Phi);\end{split}\\
&& \hphantom{\pi_{\directdefeat} = \big \{} \subformula(X) \la \checkDirectdefeat(\_,X);\\
&& \! \begin{split}\hphantom{\pi_{\directdefeat} = \big \{} & \ptrue(\directdefeat(T),X) \lor \pfalse(\directdefeat(T),X) \la \hspace{-1em} && \atom(X), \\
&&& \checkDirectdefeat(T,\_);\end{split}\\
&& \hphantom{\pi_{\directdefeat} = \big \{} \attack \la \ismodel(\directdefeat(T),X), \checkDirectdefeat(T,X). \; \big \}
\end{aspworkaround}

The program $\pi_{\attacksp}$ in combination with any attack type programs (such as $\pi_{\defeat}$, $\pi_{\directdefeat}$, ...) computes the respective attack relations.
\begin{center}
\small
$\pi_{\attacksp} = \pi_{\modelcheck} \cup \pi_{\mathrm{att}} \cup \pi_{\support} \cup \pi_{<_\key} \cup \pi_{\attsat}$
\end{center}
%This approach is quite flexible and allows for adding new attack types 

\begin{example}
Continuing our running example, we now consider the flattened input as exemplified in Example~\ref{ex:flatten} and the program $\pi_{\attacksp} \cup  \pi_{\directdefeat}$. We obtain 9 answer-sets that contain the attack information between arguments:
\begin{center}
\small
$\{\attack(4,1).\}$, $\{\attack(6,2).\}$, $\{\attack(4,3).\}$, $\{\attack(6,4).\}$, $\{\attack(3,5).\}$, $\{\attack(3,4).\}$, $\{\attack(6,3).\}$, $\{\attack(4,6).\}$, $\{\attack(3,6).\}$
\end{center}
The first answer-set, for example, represents that argument $a_4 = (\{\neg b, a \rightarrow b\},\neg a)$ attacks (directly defeats) $a_1 = (\{a\},a)$.
\end{example}

\subsection{Overall Approach at a Glance}

To sum it up, the overall process of our instantiation-based approach for generating argumentation frameworks consists of the following steps:

\begin{enumerate}
 \item A knowledge-base $\knowledgebase$ and a set $\presetclaims$ of claims are used as input.  
 \item The encoding $\pi_\arguments = \pi_{\modelcheck} \cup \pi_{\arg} \cup \pi_{\entailment} \cup \pi_{<} \cup \pi_{\mini} \cup \pi_{\consistent}$ 
defines how arguments are derived from $\knowledgebase$ and $\presetclaims$: $\pi_{\modelcheck}$ is generally used for evaluating formulae under truth assignments. Within $\pi_{\arg}$, for each argument $A = (S,C)$ a claim $ C \in \presetclaims$ and a support $ S \subseteq \knowledgebase$ is guessed. The encodings  $\pi_{\entailment}$, $\pi_{<} \cup \pi_{\mini}$ and  $\pi_{\consistent}$ guarantee that the support entails the claim, the support is subset minimal and that $S$ is a consistent set of formulae.
%$S \models C$, $\not \exists S' \subset S$ s.t. $S' \models C$ and that $S$ is a consistent set of formulae.
 \item The resulting arguments are ``flattened'' and used as input for $\pi_\attacksp$.
 \item The encoding $\pi_\attacksp = \pi_{\modelcheck} \cup \pi_{\mathrm{att}} \cup \pi_{\support} \cup \pi_{<_\key} \cup \pi_{\attsat}$ is shared by all attack types. $\pi_{\modelcheck}$ is again needed for model checking. $\pi_{\mathrm{att}}$  guesses two arguments at once. $\pi_{\support} \cup \pi_{<_\key}$ constructs a single support formula for those arguments by connecting the contained formulae by conjunction. $\pi_{\attsat}$ saturates all attack computations.
 \item Any attack encodings, such as $\pi_{\defeat}$ and $\pi_{\directdefeat}$ can be used in combination with $\pi_\attacksp$ in order to compute the respective attack relations.
\end{enumerate}

\section{Visualization of Argumentation Frameworks}
\label{sec:visual}

In order to visualize argumentation frameworks we make use of the purpose-built tool \ASPVis{}\footnote{\url{http://dbai.tuwien.ac.at/proj/argumentation/vispartix/#ARVis}}.
 \ASPVis{} is intended for the visualization of answer-sets and their relations by means of a directed graph. Each node in the graph represents an answer-set and a directed edge between two arguments represents a relation. 
  
We now describe the process of generating and visualizing argumentation frameworks by using the encodings $\pi_{\arguments}$ and $\pi_{\attacksp}$. \ASPVis{} provides a wizard that handles the respective steps:

\begin{enumerate}
\item Obtain arguments: The program $\pi_{\arguments}$ and a problem instance must be specified within \ASPVis{}. \ASPVis{} computes the arguments by invoking an ASP solver.
\item Flatten arguments: The arguments obtained in the previous step are ``flattened'', \ie a single set of facts is generated in order to be able to reason over all obtained arguments when computing the attacks. 
\item Obtain attacks: The program $\pi_{\attacksp}$ and the attack type programs are now specified. The relations between the arguments are computed. 
\item Select attack predicate: In general, \ASPVis{} accepts any binary predicate that represents answer-set relations. We define $\attack/2$ here.
\item Argumentation framework: We obtain a graph visualization consisting of arguments (vertices) and attacks (edges). 
\item Export: The obtained argumentation framework can be exported for further processing. 
\end{enumerate}
\vspace{-3em}
\begin{figure}[!h] %H
  \caption{\ASPVis: Resulting argumentation framework}
  \centering
    \includegraphics[width=0.99\textwidth]{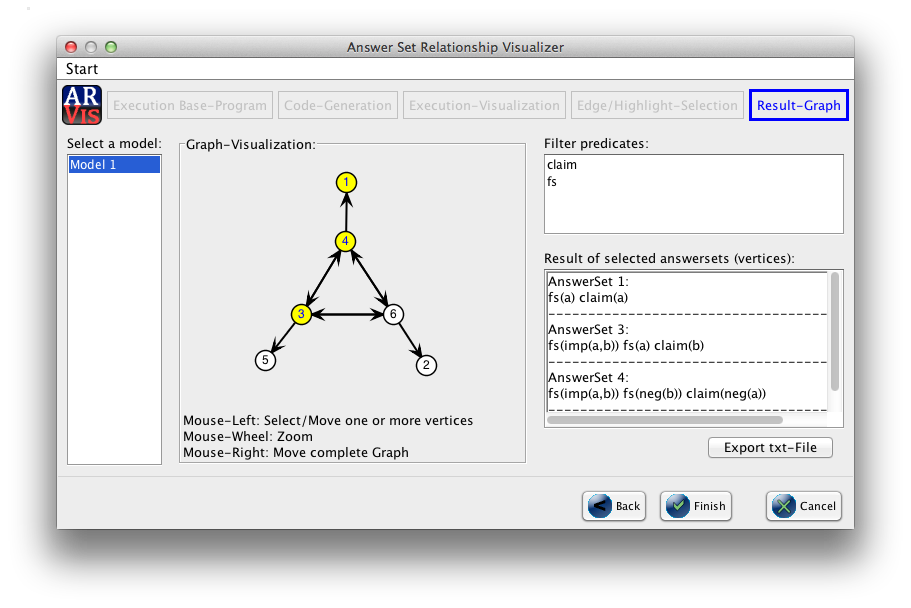}
    \label{fig:aspvis_af}
\vspace{-3em}
\end{figure}
The argumentation framework resulting from our running example, as represented by \ASPVis{}, is given in Figure~\ref{fig:aspvis_af}. The attacks correspond to direct defeats between arguments. Each argument in the graph is represented by its id. By selecting an argument its claim and support are shown in the text field on the right side. 
% Furthermore, it directly supports different ASP solvers such as dlv\footnote{\url{http://www.dlvsystem.com/}} and Clingo\footnote{\url{http://potassco.sourceforge.net/}} as well as user-specified combinations of grounder and solver. We make use of Clingo and claspD. 
All encodings, \ASPVis{} and detailed configuration information is available at
 \begin{center}
 \url{http://dbai.tuwien.ac.at/proj/argumentation/vispartix}
 \end{center}

\ASPVis{}  is a general-purpose tool that may also be used in many other areas of research: Consider, for example, the Traveling Salesperson Problem (TSP)
 where the first program computes cities and the second program outputs routes where every city is visited exactly once. In fact, \ASPVis{} can be used for any problem where one is interested in the relation between answer-sets. It is only necessary to specify two  answer-set programs: One for generating answer-sets and a second one for computing relations between those answer-sets. 
 
 Our approach is different from other available ASP visualization tools: % \todo{note: added related work from intro here}
 %\todo[inline]{currently not moved to section 4}
%There are a number of tools for the visualization of ASP. 
ASPViz \cite{CliffeVBP08} takes two answer-set programs as input, 
one for the problem encoding and one for the visualization. The latter is used for visualization for each answer-set of the former separately. It is realized in Java and works with pre-defined predicates to extract the visualization of answer-sets. 
IDPDraw \cite{Wittocx2009} works in a similar 
fashion, which augments the presentation by providing also time points to show the result in different evolutionary states. Kara \cite{KloimuellnerOPT2011} from the SeaLion development environment for ASP also provides visualization 
of answer-sets using special predicates. ASPIDE \cite{FebbraroRR11} gives the user the opportunity to visualize the dependency graph of the input program and thus allows for another type of representation. 

\section{Conclusion}
\label{sec:conclusion}

In this paper, we have presented a novel ASP-based tool for constructing 
argumentation frameworks from a given knowledge base. We have 
provided here the concrete ASP encodings 
used to obtain
such frameworks 
when logic-based arguments cf.\ \cite{BesnardH01} are employed. However, 
similar encodings for further approaches of argumentation 
are possible and subject of future 
work, %. 
%future work for performance
as well as a performance evaluation of the presented approach to check its scalability with large knowledge bases. 

When designing our tool, \ASPVis{}, we tried to keep it as 
flexible as possible such that the concrete construction of the framework can 
be specified in the logic programs.  As it has turned out, \ASPVis\ is thus
not only a tool for generating and visualizing argumentation frameworks
but also for graphically representing relations between answer-sets 
in a user-specified manner. Ongoing work thus focuses on application areas where 
it is the relation between the answer-sets (rather than the single answer-sets) 
that can support the designer of logic programs or 
where this relation is the relevant output of an ASP encoding.

%\paragraph{Summary}

%\paragraph{Future Work}

%plan to use ASP Visualizer in different domains

%WICHTIG: einschraenkung mit claims etc... future work: use cores.

\paragraph{Acknowledgments} We would like to thank Thomas Ambroz and Andreas Jusits for implementing and adapting \ASPVis{} according to our requirements as
well as Torsten Schaub for providing hints how to process arbitrary formulae 
via ASP. 

This work has been funded by the
Vienna Science and Technology Fund (WWTF) through
project ICT08-028, 
and by the Vienna University of Technology program ``Innovative Projects''.

% References
% \nocite{*}
\bibliographystyle{abbrv}
\bibliography{library}

\end{document}